\documentclass{article}
\usepackage[utf8]{inputenc}
\usepackage{iclr2025_conference,times, graphicx, float, booktabs, hyperref, listings, xcolor, url}
\usepackage{amsmath}
\usepackage{amssymb}
\usepackage{caption}
\usepackage{enumitem}

\lstdefinelanguage{json}{
    basicstyle=\ttfamily\small\color{black},
    numbers=left,
    numberstyle=\tiny\color{gray},
    stepnumber=1,
    numbersep=5pt,
    showstringspaces=false,
    breaklines=true,
    frame=single,
    framerule=0.5pt,
    rulecolor=\color{black},
    backgroundcolor=\color{white},
    literate=
     *{:}{{{\color{black}:}}}{1}
      {,}{{{\color{black},}}}{1}
      {"}{{{\color{black}"}}}{1},
    morestring=[b]",
    stringstyle=\color{red},
    identifierstyle=\color{black},
    commentstyle=\color{gray}
}

\title{Extracting Dataset Mentions in Forced Displacement and FCV Documents: A Weakly Supervised Framework with LLM-Based Label Refinement}

\author{Rafael Macalaba$^{\dagger}$, Aivin V. Solatorio$^{*}$,  \& Olivier Dupriez \\
Development Data Group \\
Office of the World Bank Group Chief Statistician \\
The World Bank \\
1818 H Street N.W., \\
Washington, 20433 \\
District of Columbia, USA \\
\texttt{\{rmacalaba, asolatorio, odupriez\}@worldbank.org} \\
\And
Patrick Michael Brock \\
World Bank UNHCR Joint Data Center \\
UN City \\
Marmorvej 51, 2100 \\
Copenhagen, Denmark
\texttt{brock@unhcr.org}
}

\iclrfinalcopy

\begin{document}

\maketitle

\begingroup
\renewcommand\thefootnote{\fnsymbol{footnote}}
\footnotetext[1]{GitHub/HF: \href{https://github.com/avsolatorio}{\texttt{@avsolatorio}}, \href{mailto:avsolatorio@gmail.com}{avsolatorio@gmail.com}}
\footnotetext[2]{GitHub: \href{https://github.com/rafmacalaba}{\texttt{@rafmacalaba}}, \href{mailto:rafael.macalaba@yahoo.com}{rafael.macalaba@yahoo.com}}
\endgroup

\begin{abstract}

Development and humanitarian organizations produce and support surveys, administrative registries, and other data resources to inform research, policy, and operations, yet systematically identifying where these datasets are referenced remains difficult. Such references are dispersed across research papers, project documents, humanitarian reports, and other unstructured text, limiting both the ability to trace data use and to identify potential gaps in data availability or dissemination. We present a weakly supervised framework for adapting dataset extraction to forced displacement and Fragile, Conflict, and Violence (FCV) documents without first constructing a large manually labeled training corpus. A lightweight model trained on general research literature generates candidate dataset mentions from unlabeled domain documents, which a frontier large language model (LLM) reviews in context, validating or rejecting candidates and correcting their extraction boundaries. The resulting annotations are supplemented with targeted synthetic and contrastive examples and used to fine-tune the lightweight model for large-scale extraction. We evaluate the resulting model on an independent gold-standard benchmark of 1,706 text passages spanning research, humanitarian, and operational documents. Across the full benchmark, the model achieves 74.1\% precision and 70.5\% recall at the mention level; among passages containing dataset references, precision reaches 89.5\%. At the passage level, the model achieves 88.2\% accuracy and 88.6\% specificity in distinguishing passages with dataset references from those without them. These results demonstrate a practical approach for constructing domain-specific supervision when labeled data are limited, and provide a technical foundation for larger-scale analysis of data use and potential gaps in the displacement data landscape.

\end{abstract}

\section{Introduction}
\label{sec:introduction}

Empirical research, policy development, and operational decision-making in Fragile, Conflict, and Violence (FCV) and forced-displacement settings can draw on household surveys, administrative registries, socioeconomic assessments, and other specialized data resources. However, systematically tracking where these resources are referenced remains difficult. Dataset citation practices are inconsistent, and references are not always captured through standardized citations or persistent identifiers \citep{buneman2021linked,piwowar2013data,silvello2018citing}.

This creates a particular opportunity for development and humanitarian organizations that produce, finance, and disseminate data in data-scarce environments. Datasets produced or supported by organizations such as the World Bank, the United Nations High Commissioner for Refugees (UNHCR), and the Joint Data Center on Forced Displacement (JDC) may appear across research papers, policy documents, project documents, humanitarian reports, footnotes, and other unstructured text rather than through consistent bibliographic citations \citep{mooney2012data}. Identifying these references at scale can serve two complementary purposes. First, it can reveal where existing data resources are being taken up in research, policy, and operations, providing evidence about their reach and contribution to evidence-informed analysis and decision-making. Second, when combined with information on what data exist, where they are available, and which populations and topics they cover, patterns of use and non-use can help identify potential gaps in data production, accessibility, or dissemination. Such evidence can therefore support not only retrospective monitoring of data investments, but also prospective decisions about where additional investment in data production and dissemination may be most valuable.


Automated information extraction offers a practical solution. Recent models
have improved the ability to identify entities and structured information from
unstructured text \citep{numind2024nuextract,hussain2023defnlp}, while
generalist named-entity recognition models such as \texttt{GLiNER}
\citep{zaratianna2023gliner} provide a flexible foundation for identifying
specialized entity types. The main challenge in applying these methods to FCV
and forced-displacement documents is the limited availability of
domain-specific labeled data. Models trained on research literature may not
transfer directly to specialized documents, while manually constructing a
large training corpus would require substantial expert annotation.

We address this problem by using an existing dataset-extraction model as a
source of weak supervision. A lightweight model previously trained to identify
dataset mentions in general research literature
\citep{macalaba2026aimonitoringclassifyingdata} is first used to generate
candidate mentions from unlabeled FCV and forced-displacement documents. A
frontier large language model (LLM) then reviews these candidates in context,
validates or rejects them, and corrects their extraction boundaries. The
resulting annotations are supplemented with targeted synthetic and contrastive
examples and used to fine-tune the lightweight model for the target domain.
This approach uses the LLM to construct higher-quality training data while
retaining a lightweight model for large-scale inference.

We evaluate the resulting model on an independent gold-standard benchmark of
1,706 text passages spanning research, humanitarian, and operational documents.
The evaluation considers both whether the model correctly extracts dataset
mentions and whether it correctly distinguishes passages containing dataset
references from those containing none.

The paper makes three contributions. First, we introduce a weakly supervised
framework that combines an existing extraction model, LLM-assisted annotation
refinement, and targeted synthetic data to construct domain-specific training
data when labeled examples are scarce. Second, we show how this supervision can
be transferred to a lightweight local model for large-scale dataset extraction.
Third, we evaluate the resulting model on an independent, heterogeneous
gold-standard benchmark covering multiple types of FCV and forced-displacement
documents.

Reliable extraction of dataset references is one technical building block for constructing a broader picture of the displacement data landscape. A textual reference does not by itself establish how a dataset was used, whether it influenced a decision or outcome, or whether the absence of a reference indicates a genuine data gap. The present study therefore evaluates the extraction capability rather than attempting to estimate the full landscape of displacement data use. Its benchmark is designed to test transfer across heterogeneous research, humanitarian, and operational document types, not to constitute a representative corpus of all policy, research, and operational materials relevant to forced displacement. Applying the framework to that broader substantive objective would additionally require systematic corpus construction, displacement-relevance filtering, dataset resolution and linkage, and contextual classification of how identified data are used. We return to these extensions in future work.


\section{Related Work}
\label{sec:related_work}

\subsection{Dataset Citation and Discoverability}

Inconsistent data citation practices are a long-standing barrier to tracking dataset use. \citet{mooney2012data} document the absence of a stable citation convention for datasets analogous to that for scholarly publications, and \citet{silvello2018citing} formalizes the requirements a data citation system must satisfy to support discovery, reuse, and credit attribution. \citet{buneman2021linked} extend this to the citation graph itself, showing how linking data citations to publication citations can improve dataset discoverability at scale. \citet{piwowar2013data} provide empirical evidence that datasets with persistent identifiers are reused and cited more often, underscoring the practical value of systematically tracking references for organizations, such as the World Bank and UNHCR, that invest heavily in data production. Our work complements this literature: rather than proposing a citation standard, we extract dataset references as they actually appear in unstructured text, which remains the dominant mode of reference in development and humanitarian documents today.

The relationship between citation practice and observed data use also motivates a potential downstream application of the extraction framework. More standardized and persistent references may make datasets easier both for users to discover and for organizations to trace across downstream documents. The present study does not test whether improved citation practice causes greater data use. However, applying the extractor at scale could enable future empirical analysis of whether datasets with stronger documentation, persistent identifiers, standardized citation guidance, or greater accessibility are subsequently referenced across a wider range of research, policy, and operational documents. Such analysis could provide an additional evidence base for investments in data documentation and dissemination.

\subsection{Automated Dataset Mention Extraction}

A growing line of work applies information extraction methods directly to the dataset-mention identification problem. \citet{heddes2021automatic} apply a transformer-based named-entity-recognition tagger to detect dataset names in scientific articles, and \citet{hussain2023defnlp} combine transformer-based question-answering and named-entity-recognition models within a similar extraction framework applied to the same underlying task. \citet{younes2023question} directly compare named-entity-recognition and question-answering approaches for extracting previously unseen dataset names, finding that question-answering formulations generalize better to datasets absent from training data -- motivating our own use of an LLM, rather than a fixed tagger, for candidate adjudication. Beyond individual research contributions, \citet{potok2022datashow} describes ``Show US the Data,'' a multi-agency U.S. government pilot applying AI methods to improve the discoverability of public datasets referenced in the literature, illustrating that automated dataset-mention tracking is also an active institutional priority and not solely a research problem. Our own prior work builds a dataset-extraction model for research literature using synthetic and LLM-assisted data \citep{solatorio2025largelanguagemodelssynthetic} and evaluates it at scale on World Bank Policy Research Working Papers \citep{macalaba2026aimonitoringclassifyingdata,macalaba2025tracking}. The present paper extends this line of work in a new direction: rather than improving extraction on research literature, we adapt an existing research-literature model to a substantially different document distribution -- humanitarian, operational, and FCV documents -- where dataset references are sparser, less standardized, and more often expressed alongside project titles and administrative systems. This domain-transfer setting is the primary methodological focus of the present work.

\subsection{Weak Supervision and LLM-Assisted Annotation}

Our candidate-generation-then-refinement design draws on the weak-supervision
paradigm formalized by \citet{ratner2017snorkel}, in which noisy, programmatic,
or model-generated labels are combined and denoised rather than manually
curated from scratch. Where Snorkel-style systems typically combine multiple
labeling functions probabilistically, we instead use a single upstream model to
generate candidates and a frontier LLM to adjudicate them individually in
context -- a design in the spirit of LLM-based data programming and
self-training approaches such as Self-Instruct \citep{wang2022self}, in which a
strong model is used to generate and filter training signal for a downstream
task. This also relates to active learning \citep{settles2009active}, in that
both approaches aim to concentrate limited annotation effort where it is most
informative; our approach substitutes LLM adjudication for a human oracle at
the candidate-review step, while still relying on a rigorously human-constructed
gold-standard set for final evaluation. This paper applies this LLM-refined
weak-supervision pipeline to the specific problem of transferring
dataset-mention extraction into humanitarian and FCV document collections.

\subsection{Data Use in Forced Displacement and FCV Settings}

The scale and persistence of forced displacement make reliable, accessible, and policy-relevant data a strategic priority for organizations operating in these settings \citep{unhcr2024globaltrends}. The World Bank--UNHCR Joint Data Center on Forced Displacement (JDC) Strategy 2024--2027 emphasizes four complementary priorities: systematic inclusion of forcibly displaced and stateless populations in national statistics; targeted production of high-quality data and timely analysis to inform policy and programs; data innovation to improve the quality, timeliness, and accessibility of data; and operationalizing data and evidence to strengthen solutions to forced displacement \citep{jdc2024strategy}. The World Bank Group's FCV Strategy 2026--2030 similarly emphasizes earlier anticipation of FCV risks, differentiated engagement across FCV settings, a stronger One World Bank Group approach centered on jobs, and strengthened operational tools and partnerships \citep{worldbank2026fcv}. UNHCR's data transformation strategy \citep{unhcr2020datastrategy} further emphasizes the role of data systems, technology, and responsible data use in humanitarian settings.

These strategic priorities complement the practical and governance challenges of producing and using data in fragile and displacement-affected settings. \citet{hoogeveen2020data} document the practical work involved in producing reliable survey data in fragile settings, while the Inter-Agency Standing Committee's operational guidance \citep{iasc2023data} sets out governance principles for humanitarian data. Beyond data production itself, however, organizations that produce, fund, and disseminate these resources also stand to benefit from systematic visibility into where and how the resulting data are subsequently referenced.

Such visibility has potential value beyond accountability or retrospective monitoring. Evidence on where displacement-relevant data are referenced can help characterize the reach of existing data resources across research, policy, and operational processes. When combined with information on what data are available and the populations and topics they cover, patterns of observed use and non-use may also help identify potential gaps in data production, accessibility, documentation, or dissemination. This paper develops one technical component needed for such analysis: scalable extraction of dataset references from heterogeneous FCV and forced-displacement documents.

\section{Methodology}
\label{sec:methodology}

Our goal is to adapt dataset extraction to FCV and forced-displacement
documents when domain-specific labeled training data are scarce. Rather than
first constructing a large manually annotated training corpus, we use an
existing dataset-extraction model to generate weak annotations from unlabeled
documents and a large language model (LLM) to refine them. We supplement these
annotations with targeted synthetic examples and use the combined data to
fine-tune a lightweight extraction model for the target domain.

The framework consists of four stages. First, a lightweight model previously
trained to extract dataset mentions from research literature identifies
candidate mentions in unlabeled FCV and forced-displacement documents. Second,
a frontier LLM reviews these candidates in their original context, removes
invalid predictions, and corrects extraction boundaries. Third, targeted
synthetic and contrastive examples expand the training data beyond the cases
observed in the available documents. Finally, the annotations from real
documents and the synthetic examples are combined to fine-tune the lightweight
model for large-scale extraction.

This design separates \textit{training-data construction} from
\textit{inference}. The existing extractor provides an efficient source of weak
supervision, while the LLM is used selectively to improve annotation quality.
Synthetic generation expands coverage of difficult or underrepresented cases.
The resulting supervision is then transferred to a lightweight local model that
can be applied independently at scale, without requiring LLM verification at
inference time.

\subsection{Weakly Supervised Candidate Generation}
\label{sec:candidate_generation}

The first stage identifies possible dataset references in previously unlabeled
FCV and forced-displacement documents. Rather than manually reviewing every
passage, we use an existing extraction model to locate text spans that may refer
to datasets. These predictions serve as \textit{weak annotations}: they reduce
the annotation search space but are not assumed to be correct.

PDF documents are processed page by page using a standardized text-extraction
utility. The extracted text is divided into page-level chunks and passed to
\texttt{GLiNER2} (\texttt{ai4data/datause-extraction}), a lightweight local
model previously trained for dataset-mention extraction using general research
publications and World Bank Policy Research Working Papers
\citep{zaratianna2023gliner,macalaba2026aimonitoringclassifyingdata}. The model
therefore provides an initial dataset-extraction capability but has not yet been
adapted to the FCV and forced-displacement documents considered here.

At this stage, the objective is to maximize candidate coverage rather than
precision. We therefore apply a permissive prediction threshold of
$\tau=0.15$, retaining uncertain predictions that would normally be discarded
at a higher inference threshold. This produces an intentionally broad candidate
set, which the LLM-assisted refinement stage then narrows and corrects.

For each candidate, the model returns the predicted text span, character
offsets, and initial classification. These predictions are retained together
with their surrounding source text and passed to the LLM-assisted refinement
stage. The local model therefore acts as a screening mechanism rather than a
final annotator: it identifies where potential dataset references occur, while
the subsequent stage determines whether each candidate should be retained,
corrected, or discarded.

\subsection{LLM-Assisted Annotation Refinement}
\label{sec:llm_refinement}

The weak annotations generated in the first stage are useful for locating
potential dataset references, but benefit from a second, context-aware pass
before serving as domain-specific training data. We therefore use
\texttt{GPT-4o} as a context-aware annotation judge. Each candidate is
evaluated together with its surrounding source text according to a predefined
annotation protocol.

The refinement stage performs two functions that are directly relevant to the
extraction task:

\begin{enumerate}
    \item \textbf{Candidate validation}: determine whether the predicted span
    refers to a structured data resource, such as a survey, database,
    administrative registry, census, microdata collection, or other
    identifiable dataset; and

    \item \textbf{Boundary correction}: adjust the predicted span when the
    initial extraction contains too much or too little text.
\end{enumerate}

Candidates referring instead to projects, organizations, analytical methods,
narrative publications, qualitative methods, document headings, or
administrative tools that do not themselves constitute data resources are
excluded.

For example, consider the following passage from the World Bank's
\textit{Jordan Emergency Food Security Project} Project Appraisal Document:

\begin{quote}
\small
\textit{``Existing food insecurity levels are particularly high among Jordan's
refugee population. According to the most recent mobile Vulnerability Assessment
and Mapping (mVAM) completed by the World Food Program (WFP) in Jordan, 7
percent of Jordanian households were found to be food insecure as of February
2021...''}
\end{quote}

The first-stage model identifies \texttt{mobile Vulnerability Assessment and
Mapping (mVAM)} as a candidate dataset reference. The LLM retains the candidate
because the surrounding text indicates that mVAM refers to a structured data
resource used to measure food insecurity.

In contrast, consider the following passage from the UNHCR
\textit{Mindanao Forced Displacement Annual Report}:

\begin{quote}
\small
\textit{``The Annual Mindanao Displacement Dashboard aims to provide a starting
point for information and analysis that can help protection agencies, policy
makers and other stakeholders concerning instances of forced displacement...''}
\end{quote}

At the permissive candidate-generation threshold, the first-stage model
identifies \texttt{Annual Mindanao Displacement Dashboard} as a possible
dataset reference. The LLM rejects the candidate because the surrounding text
indicates that it is an information product rather than an underlying
structured dataset.

The LLM response is constrained through OpenAI's structured output interface
using a Pydantic schema. This ensures that validation decisions, corrected
spans, and associated annotations are returned in a consistent machine-readable
format. The complete response schema, system prompt, and inclusion and exclusion
rules are provided in Appendix~\ref{app:llm_refinement}.

This stage converts an initial, broad set of model predictions into a cleaner
set of domain-specific annotations without requiring the LLM to search every
passage in the document collection from scratch. Because the refinement stage
operates on candidates surfaced in the first stage, the overall pipeline's
recall depends on the coverage of the initial candidate generator; we discuss
this design choice further in Section~\ref{sec:discussion}.

\subsection{Annotation Schema}
\label{sec:annotation_schema}

Dataset references vary substantially in how precisely they identify the
underlying data resource. To provide a consistent annotation target, validated
mentions are classified into three levels of specificity:

\begin{itemize}
    \item \texttt{named}: a formally named survey, database, registry, or other
    data resource with a recognizable title or acronym, such as the
    \textit{Demographic and Health Survey}, \textit{LSMS}, or
    \textit{World Development Indicators};

    \item \texttt{descriptive}: a specific description of a data resource that
    does not use a unique formal name, such as \textit{survey of Javanese farm
    households}; and

    \item \texttt{vague}: a general reference to a data resource without enough
    information to identify a particular dataset, such as
    \textit{administrative data} or \textit{utility records}.
\end{itemize}

The annotation schema also supports additional metadata fields, including the
dataset acronym, publisher, geography, reference population, and publication
year when these are available in the surrounding text. These fields provide
structured information about the identity and scope of a referenced data
resource, but they are not part of the extraction evaluation reported in this
paper.

The full annotation schema and validation rules are provided in
Appendix~\ref{app:llm_refinement}.

\subsection{Targeted Synthetic Data Generation}
\label{sec:synthetic_generation}

The LLM-refined annotations provide domain-specific examples drawn from real
documents, but their coverage is naturally bounded by both the available
document collection and the candidates surfaced by the initial model. We
therefore supplement these annotations with synthetic examples. The purpose of
synthetic generation is not to reproduce the distribution of the underlying
corpus, but to broaden the training data with additional dataset types,
linguistic forms, and underrepresented cases.

We use few-shot prompting with \texttt{GPT-4o-mini} to generate passages
containing dataset references. Each prompt provides annotated examples that
demonstrate the expected extraction format and is guided by selected dataset
names, dataset types, and contextual patterns for which additional examples are
desired. These include FCV-relevant resources such as the
\textit{Socio-Economic Insights Survey (SEIS)}, \textit{Multi-Sector Needs
Assessment (MSNA)}, and WFP's \textit{mobile Vulnerability Assessment and
Mapping (mVAM)}, as well as generic expressions such as \textit{rainfall data}.

We use more than 100 prompt templates spanning Development Economics,
Humanitarian and Protection, Global Health, Food Security and Agriculture,
Education, and Climate and Environment. Varying both the dataset seeds and the
contexts in which they appear increases the diversity of the generated
examples while retaining control over their annotations.

Generated records are constrained using a Pydantic schema and subjected to
automated validation before inclusion in the training corpus. The generated
text must contain the specified dataset and metadata attributes verbatim;
dataset spans cannot include standalone publication years; and spans cannot
terminate in incomplete grammatical fragments such as trailing prepositions or
verbs. Examples that fail these checks are discarded. Additional generation
details and the domain-specific variable banks are reported in
Appendix~\ref{app:synthetic_generation}.

\subsubsection{Contrastive Examples}
\label{sec:contrastive_examples}

Synthetic generation is also used to create contrastive examples. The purpose
of these examples is to expose the model to cases in which similar expressions
should receive different extraction labels depending on their context.

For example, \textit{LSMS} should be extracted when it refers to the Living
Standards Measurement Survey used as a data resource, but not when
\textit{LSMS team} refers to the people conducting fieldwork. Similarly,
\textit{household consumption survey} may identify a data resource, whereas
\textit{we conducted household surveys} describes a data-collection activity
rather than the resulting dataset. Table~\ref{tab:contrast_pairs} illustrates
these cases.

\begin{table}[h]
\centering
\caption{Illustrative contrastive synthetic examples.}
\label{tab:contrast_pairs}
\small
\begin{tabular}{p{0.18\textwidth} p{0.40\textwidth}
p{0.20\textwidth} p{0.10\textwidth}}
\toprule
\textbf{Context Type} & \textbf{Example Sentence} &
\textbf{Target Span} & \textbf{Label} \\
\midrule

\textbf{Dataset mention} &
The analysis incorporates the \textbf{Living Standards Measurement Survey
(LSMS)} to calculate consumption aggregates. &
Living Standards Measurement Survey &
\texttt{named} \\

\textbf{Non-dataset use} &
The \textbf{LSMS} team coordinated the field operations and trained the local
enumerators. &
\textit{None} &
\textit{Empty} \\

\midrule

\textbf{Dataset mention} &
We gathered the \textbf{household consumption survey} data across three
displacement camps. &
household consumption survey &
\texttt{descriptive} \\

\textbf{Non-dataset use} &
We conducted \textbf{household surveys} to assess local needs and immediate
protection risks. &
\textit{None} &
\textit{Empty} \\

\bottomrule
\end{tabular}
\end{table}

These examples encourage the model to use surrounding context rather than
relying only on familiar dataset names, acronyms, or data-related expressions.

\subsection{Fine-Tuning and Final Extraction}
\label{sec:fine_tuning}

The final training corpus combines two sources of supervision: annotations
derived from real FCV and forced-displacement documents through weakly
supervised candidate generation and LLM-assisted refinement, and targeted
synthetic examples generated through few-shot prompting. The combined corpus
is used to fine-tune \texttt{GLiNER2} for dataset extraction in the target
domain.

This final step transfers the supervision produced during training-data
construction into the lightweight extraction model. The frontier LLM is
therefore not part of the deployed extraction pipeline: once fine-tuned, the
local model can process new document collections independently.

For final extraction, the fine-tuned model is applied at a confidence threshold
of $\tau=0.40$. Its performance is evaluated against a separate gold-standard
holdout set of 1,706 text chunks. The holdout data are not used to construct
the weak annotations, refine candidates, generate synthetic examples, or
fine-tune the model. Section~\ref{sec:experimental_setup} describes the
evaluation corpus and benchmark construction in detail.

\section{Data and Evaluation Design}
\label{sec:experimental_setup}

The document collections in this study constitute a technical validation corpus, rather than a representative corpus from which to characterize the overall landscape of displacement data use. Documents were selected to expose the extraction framework to heterogeneous settings in which dataset references differ in frequency, form, and surrounding terminology. The resulting evaluation therefore addresses whether the extraction method transfers across document types; it does not estimate the prevalence or distribution of data use across research, policy, and operations.

A full application to mapping the displacement data landscape would require a separately designed corpus with explicit coverage of the relevant evidence ecosystem—for example, academic research, government strategies and policy documents, humanitarian assessments, and development operations—as well as a mechanism for identifying displacement-relevant passages within documents whose overall subject may be broader. These corpus-construction and relevance-classification tasks are outside the scope of the present technical evaluation.


\subsection{Document Collections}

The corpus comprises 93 documents purposively assembled from four document collections chosen to provide variation in document format, substantive context, and the explicitness with which datasets are referenced. The collections span settings ranging from formal research publications, where datasets are often identified explicitly, to humanitarian and operational documents, where references may be more implicit or embedded in narrative, administrative, and monitoring language. The resulting composition is intended to expose the extraction framework to heterogeneous forms of dataset mention rather than to constitute a representative sample of the broader forced-displacement and FCV evidence ecosystem.

\begin{itemize}

\item \textbf{UNHCR/ReliefWeb reports (43 documents):}
Humanitarian reports and field-level updates covering displacement, protection needs, and conditions affecting displaced populations. This collection provides a relatively heterogeneous reporting environment in which references to surveys, assessments, registries, and other data resources may be sparse, implicit, or expressed without formal citation conventions. A larger number of documents was therefore included to obtain sufficient coverage of these less standardized forms of dataset reference.

\item \textbf{SEIS and humanitarian briefs (25 documents):}
Socio-Economic Insights Surveys (SEIS) and related analytical briefs focused on the socioeconomic conditions of displaced populations. These documents provide an analytically oriented setting in which datasets and empirical findings are more central to the text than in field-level humanitarian reporting, while remaining less standardized than formal research publications. The collection also includes documents identified by subject-matter experts as thematically relevant to forced displacement.

\item \textbf{World Bank Policy Research Working Papers (14 documents):}
Research papers in which datasets are typically identified explicitly as part of empirical analysis. This collection serves as the most citation-explicit document type in the benchmark and is also closest to the source domain of the initial extraction model, which was trained primarily on research-oriented documents of this kind. It therefore provides a useful comparison for assessing how performance changes as the model is applied to less structured and less citation-standardized domains.

\item \textbf{World Bank Project Appraisal Documents (11 documents):}
Operational documents containing narrative text, indicators, tables, project and administrative terminology, and monitoring frameworks. These documents represent a high-ambiguity setting in which project titles, administrative systems, reports, monitoring platforms, and other information products may resemble dataset references. Their inclusion is particularly important for evaluating whether the model can distinguish genuine data sources from dataset-adjacent entities in operational text. More broadly, project documents are relevant to the downstream objective of examining where and how data resources are referenced in the design and implementation of development operations.

\end{itemize}

Candidates were chunked, pre-extracted, LLM-judged, then human-adjudicated
with agreement-based filtering (conflicts reviewed manually), and splits
are document-disjoint to prevent leakage.

\subsection{Gold-Standard Evaluation Set}

The final model is evaluated against a manually annotated holdout benchmark of 1,706
text chunks. Of these, 545 contain at least one annotated dataset reference and
1,161 contain no dataset reference. The large share of negative passages
reflects an important feature of large-scale extraction: most document passages
do not contain dataset references, so the model must learn both when to extract
and when to return nothing. Table~\ref{tab:evaluation_distribution} summarizes the benchmark.

\begin{table}[h]
\centering
\caption{Composition of the gold-standard holdout benchmark.}
\label{tab:evaluation_distribution}
\small
\begin{tabular}{lrrr}
\toprule
\textbf{Document Collection} &
\textbf{Total Chunks} &
\textbf{With Dataset} &
\textbf{Without Dataset} \\
\midrule
World Bank PADs & 129 & 7 & 122 \\
World Bank PRWPs & 259 & 67 & 192 \\
UNHCR / ReliefWeb & 169 & 22 & 147 \\
SEIS / Baseline Holdout & 1,149 & 449 & 700 \\
\midrule
\textbf{Total} & \textbf{1,706} & \textbf{545} & \textbf{1,161} \\
\bottomrule
\end{tabular}
\end{table}

Dataset prevalence varies substantially across the benchmark. Approximately
39\% of SEIS/baseline passages contain a dataset reference, compared with about
5\% of PAD passages, allowing us to evaluate the model under both data-rich and
data-sparse conditions.

\subsection{Evaluation Metrics}

We evaluate the final model at a prediction confidence threshold of $\tau=0.40$. Because the task involves identifying both the presence and the exact textual location of dataset references, we evaluate performance at two levels: dataset-mention extraction and passage-level classification.

\subsubsection{Dataset-Mention Extraction}

At the mention level, a prediction is considered correct when it sufficiently overlaps with an annotated dataset reference. We measure overlap using the Jaccard similarity between the sets of tokens in the predicted span, $S_1$, and the annotated span, $S_2$:

\begin{equation}
J(S_1,S_2) = \frac{|W_1 \cap W_2|}{|W_1| + |W_2| - |W_1 \cap W_2|},
\end{equation}

where $W_1$ and $W_2$ are the sets of tokens contained in the predicted and annotated spans, respectively. We consider a prediction to match the annotation when $J(S_1,S_2) \geq 0.5$.

We report precision and recall. Precision measures the proportion of predicted dataset mentions that correspond to valid annotated references, while recall measures the proportion of annotated dataset mentions recovered by the model. We also report the $F_{0.5}$-score, which gives greater weight to precision than recall. For downstream landscape analysis, false-positive references could distort estimates of both observed data use and apparent gaps in use, making precision particularly important.

\subsubsection{Performance on Passages With and Without Dataset Mentions}

We report mention-level performance under two complementary settings. First, we evaluate the model across the complete holdout benchmark, including passages both with and without dataset references. This represents the model's expected behavior when processing complete document collections.

Second, we report extraction performance on passages that contain at least one annotated dataset reference. This isolates the model's ability to identify datasets when they are actually present. Comparing the two settings helps distinguish errors in dataset identification from false predictions generated on passages that contain no datasets.

\subsubsection{Passage-Level Classification}

Finally, we evaluate whether the model can correctly determine when a passage contains any dataset reference. For this analysis, each text chunk is treated as a binary classification problem. A passage is predicted as positive if the model extracts at least one dataset mention and negative if it extracts none.

We report accuracy, recall, $F_1$-score, and specificity. Specificity, or the true negative rate, measures the proportion of passages without dataset references for which the model correctly returns no extraction. This metric is particularly important for operational documents, where dataset mentions are sparse and most text consists of administrative or project-related content.

Together, the mention-level and passage-level evaluations capture two requirements of the intended system: accurately identifying dataset references when they occur and avoiding spurious extractions when they do not.

\section{Results}
\label{sec:results}

We evaluate the model produced by the semi-supervised framework on two
complementary capabilities. First, we assess whether it can correctly extract
dataset mentions and their textual boundaries. Second, we assess whether it can
distinguish passages that contain dataset references from those that do not.
Together, these evaluations capture the two capabilities required for
large-scale dataset detection: finding dataset references when they occur and
avoiding spurious extractions when they do not.

\subsection{Dataset Mention Extraction}

Table~\ref{tab:span_results} reports mention-level extraction performance on the
1,706-chunk gold-standard holdout benchmark at a confidence threshold of
$\tau=0.40$. Across the full benchmark, the model correctly identifies 620
dataset mentions, producing an overall precision of 74.1\% and recall of
70.5\%.

When the evaluation is restricted to passages containing at least one annotated
dataset reference, precision increases to 89.5\%, while recall remains 70.5\%.
This result indicates that, once operating within passages where datasets are
present, nearly nine out of ten extracted mentions correspond to an annotated
dataset reference.

\begin{table}[h]
\centering
\caption{Dataset mention extraction performance at a confidence threshold of
$\tau=0.40$.}
\label{tab:span_results}
\small
\begin{tabular}{p{0.24\textwidth} p{0.14\textwidth} p{0.18\textwidth}
r r r r r}
\toprule
\textbf{Evaluation Corpus} &
\textbf{Chunks (Pos/Neg)} &
\textbf{Evaluation Set} &
\textbf{TP} &
\textbf{FP} &
\textbf{FN} &
\textbf{Precision} &
\textbf{Recall} \\
\midrule

World Bank PADs &
129 (7 / 122) &
All Chunks &
5 & 11 & 3 & 31.2\% & 62.5\% \\

& &
Positive Only &
5 & 0 & 3 & 100.0\% & 62.5\% \\

\midrule

World Bank PRWPs &
259 (67 / 192) &
All Chunks &
68 & 25 & 37 & 73.1\% & 64.8\% \\

& &
Positive Only &
68 & 9 & 37 & 88.3\% & 64.8\% \\

\midrule

UNHCR / ReliefWeb &
169 (22 / 147) &
All Chunks &
23 & 14 & 13 & 62.2\% & 63.9\% \\

& &
Positive Only &
23 & 3 & 13 & 88.5\% & 63.9\% \\

\midrule

SEIS / Baseline Holdout &
1,149 (449 / 700) &
All Chunks &
524 & 167 & 207 & 75.8\% & 71.7\% \\

& &
Positive Only &
524 & 61 & 207 & 89.6\% & 71.7\% \\

\midrule

\textbf{Overall} &
\textbf{1,706 (545 / 1,161)} &
\textbf{All Chunks} &
\textbf{620} &
\textbf{217} &
\textbf{260} &
\textbf{74.1\%} &
\textbf{70.5\%} \\

& &
\textbf{Positive Only} &
\textbf{620} &
\textbf{73} &
\textbf{260} &
\textbf{89.5\%} &
\textbf{70.5\%} \\

\bottomrule
\end{tabular}
\end{table}

The positive-only results are notably consistent across document types.
Precision reaches 88.3\% for World Bank Policy Research Working Papers,
88.5\% for UNHCR/ReliefWeb documents, and 89.6\% for the SEIS/baseline
collection. The PAD subset yields 100\% precision on positive passages,
reflecting strong performance on the small number of annotated mentions
available for that collection. Recall ranges from 62.5\% to 71.7\% across the
four collections.

These results demonstrate that the domain-adapted model extracts dataset
references with strong precision across substantially different document
types. The gap between the all-chunk and positive-only results reflects the
separate task of determining when no dataset reference is present at all,
which we examine next as a standalone detection problem.

\subsection{Dataset Detection}

We therefore evaluate the model separately as a passage-level detector. A
passage is considered positive when the model extracts at least one dataset
reference and negative when it returns no extraction.

Across the full benchmark, the model correctly identifies 475 of the 545
passages containing dataset references and correctly returns no extraction for
1,029 of the 1,161 passages without dataset references. This corresponds to
an overall detection accuracy of 88.16\%, recall of 87.16\%, and specificity
of 88.63\%.

\begin{table}[h]
\centering
\caption{Passage-level dataset detection performance.}
\label{tab:binary_results}
\small
\begin{tabular}{l c p{0.28\textwidth} c c c}
\toprule
\textbf{Document Collection} &
\textbf{Chunks} &
\textbf{TP / FP / FN / TN} &
\textbf{Accuracy} &
\textbf{$F_1$} &
\textbf{Specificity} \\
\midrule

World Bank PRWPs &
259 &
57 / 15 / 10 / 177 &
90.3\% &
82.0\% &
92.2\% \\

UNHCR / ReliefWeb &
169 &
19 / 11 / 3 / 136 &
91.7\% &
73.1\% &
92.5\% \\

World Bank PADs &
129 &
4 / 11 / 3 / 111 &
89.1\% &
36.4\% &
91.0\% \\

SEIS / Baseline Holdout &
1,149 &
395 / 95 / 54 / 605 &
87.0\% &
84.1\% &
86.4\% \\

\bottomrule
\end{tabular}
\end{table}

Detection performance is stable across the four document collections.
Accuracy ranges from 87.0\% to 91.7\%, while specificity ranges from 86.4\% to
92.5\%. This is a meaningful result given that 1,161 of the 1,706 passages in
the benchmark contain no dataset reference at all: the model performs well not
only in finding passages that contain datasets, but also in filtering out the
much larger set of passages where no extraction should be made.

The PAD results merit a different interpretation because of the strong class
imbalance in this subset. Only 7 of 129 PAD passages contain an annotated
dataset reference. The model correctly identifies four of these passages and
correctly rejects 111 of the 122 negative passages, producing high accuracy
(89.1\%) and specificity (91.0\%); the $F_1$-score (36.4\%) is more sensitive to
the small absolute number of positive examples in this subset than to the
model's underlying behavior. This illustrates the value of reporting
specificity and accuracy alongside $F_1$ when evaluating rare-event extraction
in operational documents.

\subsection{Performance Across Document Types}

The holdout benchmark deliberately spans research, humanitarian, and operational documents that differ substantially in structure, writing style, and the ways in which dataset references are expressed. This heterogeneity provides a test of whether the model has learned dataset-reference patterns that generalize across document types rather than patterns specific to a single collection.

The clearest evidence of cross-document performance comes from extraction precision on passages containing dataset references. Despite substantial differences across collections, precision remains close to 90\% for PRWPs, UNHCR/ReliefWeb documents, and the SEIS/baseline collection. Passage-level specificity is similarly stable, exceeding 86\% in every collection. Together, these results suggest that the model transfers effectively across the research and humanitarian document types represented in the benchmark while retaining a strong ability to reject passages that do not contain dataset references.

The diversity of the benchmark also provides some evidence that the learned extraction capability is not tied to a single document genre. However, the benchmark does not include the broader range of government strategies, national policy documents, and other policy-oriented materials that would be needed to establish transfer to policy corpora. The present results therefore support the plausibility of such transfer, but direct evaluation on these document types remains necessary.

Operational documents present a particularly demanding setting because they contain many named entities and expressions that resemble dataset references without actually referring to datasets. To better characterize this challenge, we examine candidates generated from PADs during the weak-supervision stage. At the permissive candidate-generation threshold of $\tau=0.15$, 145 of 203 candidates are rejected during LLM-assisted refinement. Project and program titles constitute the largest category of rejected candidates, accounting for 50.3\%, followed by administrative software and monitoring systems at 17.2\%. Other rejected candidates include publications and standardized document headings.

This analysis identifies a recurring source of ambiguity in operational documents: distinguishing genuine dataset references from dataset-adjacent entities such as project titles, administrative systems, publications, and standardized headings. The contrastive training examples described in Section~\ref{sec:contrastive_examples} are designed specifically to expose the model to these difficult negative cases. The results therefore highlight both the importance of negative-example construction for domain adaptation and a concrete area in which further expansion of the training data may improve performance on operational documents.

\subsection{Summary of Evaluation Results}

Overall, the evaluation provides evidence that the semi-supervised framework
can adapt a general dataset-extraction model to specialized FCV and
forced-displacement documents without requiring a large manually labeled
training corpus. The resulting model achieves 89.5\% precision when extracting
mentions from passages containing datasets and identifies whether a passage
contains a dataset with 88.16\% accuracy across the complete holdout benchmark.

These results also point toward clear next steps. Mention-level recall of
approximately 70\% suggests room to expand candidate coverage in the
weak-supervision stage, and data-sparse operational documents remain the
setting most likely to benefit from additional contrastive training examples
that sharpen the boundary between datasets and project titles or administrative
systems. Together, the results demonstrate that the framework produces a
practical, deployable dataset detector from limited domain-specific
supervision, with a clear path to further gains.

\section{Discussion}
\label{sec:discussion}

This study shows that a general dataset-extraction model can be adapted to
specialized FCV and forced-displacement documents without first constructing a
large manually labeled training corpus. By combining weakly supervised candidate
generation, LLM-assisted annotation refinement, and targeted synthetic data, the
framework produces a domain-adapted local model that performs consistently
across research, humanitarian, and operational documents.

The results are particularly strong when dataset references are present.
Precision reaches 89.5\% on positive passages and remains close to 90\% across
the major document collections. At the same time, evaluation on the complete
benchmark shows why extraction performance should be assessed on more than just
passages known to contain datasets. Most passages contain no dataset reference,
making the ability to return no extraction a central part of the task. The
model's 88.6\% specificity indicates that it distinguishes dataset-bearing
passages from other content well, and the framework provides a clear structure
-- contrastive training examples -- for continuing to improve on this.

\subsection{Adapting Extraction Models with Limited Labeled Data}

A central contribution of the framework is how it constructs domain-specific
supervision. Rather than treating predictions from the general-domain model as
final labels, we use them as weak annotations that identify likely dataset
references. A frontier LLM then reviews these candidates in context, removes
invalid predictions, corrects extraction boundaries, and applies a consistent
annotation schema. Synthetic generation provides additional examples for
difficult or underrepresented cases. The resulting annotations are then used to
fine-tune a lightweight model for large-scale extraction.

This division of labor allows each model to serve a different purpose: the
general model provides broad candidate generation, the LLM provides
context-sensitive refinement, and the fine-tuned local model provides scalable
inference. The approach is particularly valuable when manually annotating a
large domain-specific corpus would be costly, but an existing model already
provides a reasonable starting point.

Importantly, the framework does not eliminate the need for human judgment.
Instead, it shifts expert effort away from exhaustive passage-by-passage
annotation toward defining annotation rules, identifying important error cases,
and constructing a rigorous gold-standard evaluation set -- a more scalable use
of scarce domain expertise.

\subsection{Learning What Not to Extract}

The evaluation also highlights a distinctive feature of this domain: learning
what constitutes a dataset is only part of the task. The model must also learn
to distinguish datasets from other entities that look similar.

This is most visible in operational documents. Project titles, administrative
systems, monitoring tools, publications, and standardized headings can share
many surface characteristics with dataset names. The PAD analysis in
Section~\ref{sec:results} shows that these ``near-miss'' entities account for a
substantial share of filtered candidates during LLM-assisted refinement.

This suggests that continued gains may come as much from expanding negative and
contrastive examples as from adding more positive dataset mentions. In
specialized document collections, explicitly teaching the model the boundary
between the target entity and plausible alternatives is a particularly
effective lever. This insight directly motivated the contrastive synthetic
examples used in our training procedure and is a natural focus for continued
development of the framework.

\subsection{From Dataset Detection to Mapping the Displacement Data Landscape}

The value of dataset-mention extraction extends beyond counting references. When extracted mentions are linked to identifiable data resources and interpreted alongside information on data availability and substantive information needs, patterns of observed use and non-use can contribute to a broader picture of the displacement data landscape. Repeated references can provide evidence that particular data resources are reaching downstream research, policy, or operational processes. Conversely, limited observed use of an existing and relevant dataset may warrant investigation of barriers related to accessibility, documentation, discoverability, timeliness, or dissemination. Where important information needs are not adequately served by existing data resources, the same landscape analysis may instead point to potential gaps in data production.

Such interpretations cannot be made from dataset mentions alone. The absence of an observed reference does not establish that a dataset was unavailable, unused, or insufficient, particularly when the underlying document corpus is not designed to be representative. Moving from mention extraction to substantive landscape analysis would therefore require additional components: systematic corpus construction spanning research, policy, humanitarian, and operational materials; classification of displacement relevance at the document or passage level; resolution and linkage of extracted mentions to identifiable datasets; contextual classification of how those datasets are used; and comparison with inventories of available data and relevant information needs.

An important distinction in this broader framework is between the relevance of a dataset and the relevance of its use. A national household survey, for example, may include displaced populations but be referenced in a policy document only for national-level findings unrelated to displacement. Conversely, the same survey may be used specifically to analyze outcomes among refugees or internally displaced populations. Establishing displacement relevance therefore requires examining the context in which a dataset is referenced, rather than inferring relevance from the identity or coverage of the dataset alone.

The present study establishes one technical component of this broader system: scalable identification of dataset references across heterogeneous documents. Combined with the additional components described above, this capability could contribute to systematic analysis of where displacement-relevant data are reaching research, policy, and operational processes, where existing data resources may face barriers to uptake, and where unmet information needs may motivate additional data production.

\subsection{Future Directions}

Several directions could further improve and validate the extraction framework itself. The training procedure combines LLM-refined annotations from real documents with synthetic and contrastive examples. The consistent precision observed on positive passages across document types (Section~\ref{sec:results}), together with the candidate-filtering analysis in Section~\ref{app:rejected_candidates}, suggests that exposure to difficult negative cases may be particularly important for distinguishing genuine dataset references from project titles, administrative systems, and other dataset-adjacent entities in operational documents. Targeted ablation studies could quantify the contribution of each source of supervision. Expanding candidate coverage during weak supervision, for example through an additional candidate-generation model or method, could help improve mention-level recall. Measuring agreement between LLM-assisted and expert annotations on a held-out sample would provide further evidence on the reliability of the refinement stage, while expanding the operational-document benchmark would strengthen evaluation in settings where dataset references are relatively sparse and ambiguous.

A second direction is to evaluate how well the extraction capability transfers beyond the document types represented in the current benchmark. In particular, direct evaluation on government strategies, national policy documents, and other policy-oriented corpora would be necessary before making broader claims about performance across the displacement evidence ecosystem. Such evaluation could also test document- and passage-level methods for identifying displacement relevance, including cases in which broader national datasets are used specifically to produce evidence about displaced populations.

The principal longer-term direction is to integrate these capabilities into a broader displacement-data landscape pipeline. This would entail constructing a systematic application corpus, resolving extracted mentions to identifiable datasets, classifying the context and substantive role of each reference, and linking observed references to inventories of available data. Such an application could then examine relationships between data availability and observed uptake and investigate whether apparent gaps are more consistent with limitations in data production or with barriers related to accessibility, documentation, discoverability, timeliness, or dissemination. These extensions would move the framework from identifying dataset references toward a more systematic evidence base for monitoring the reach of existing data investments and informing future decisions about data production and dissemination.

\section{Conclusion}
\label{sec:conclusion}

Tracking dataset references across large document collections is difficult when citation practices are inconsistent and domain-specific labeled data are scarce. This paper presents a weakly supervised framework for adapting dataset extraction to FCV and forced-displacement documents without requiring a large manually annotated training corpus. The approach combines candidate generation from an existing extraction model, LLM-assisted annotation refinement, targeted synthetic and contrastive examples, and fine-tuning of a lightweight local extraction model.

Evaluation on an independent gold-standard benchmark demonstrates that the resulting model can identify dataset references across heterogeneous research, humanitarian, and operational documents while maintaining strong precision on passages containing dataset mentions. The results also highlight a distinctive challenge of domain adaptation in this setting: dataset references must be distinguished from semantically similar entities such as project titles, administrative systems, publications, and other information products. The candidate-refinement analysis suggests that exposure to such difficult negative cases is an important component of the adaptation process and motivates further evaluation through targeted ablation studies.

More broadly, reliable dataset extraction provides a technical foundation for examining the displacement data landscape at a scale that would be difficult to achieve through manual review alone. The present study establishes one component of that broader capability: scalable identification of dataset references across heterogeneous documents. Combined with systematic corpus construction, displacement-relevance classification, dataset resolution and linkage, contextual classification of how data are used, and information on available data resources, this capability could support analysis of where displacement-relevant data are reaching research, policy, and operational processes, where existing resources may face barriers to uptake, and where unmet information needs may indicate potential gaps in data production. Such analysis would extend the value of dataset-reference extraction beyond monitoring the downstream reach of existing data investments toward informing future decisions about data production, accessibility, documentation, and dissemination.

\section*{Open Source, Ethics, Disclosure, and Disclaimer}

\textbf{Code and Model Availability.} The fine-tuned extraction model, model card, inference examples, and supporting documentation are publicly available at https://huggingface.co/collections/ai4data/data-use-displacement.

\textbf{Ethics Statement.}
This study analyzes publicly available research papers, project documents, and
humanitarian reports. It does not process individual-level records or personally
identifiable information about displaced or vulnerable populations; all analyses
are conducted at the level of document text. The framework is intended to help
data-producing organizations systematically identify where surveys,
administrative data, and other data resources are referenced across research,
policy, and operational documents. Because automated extraction is imperfect,
human review remains appropriate when model outputs are used for consequential
monitoring, reporting, or evaluation.

\textbf{Disclosure of AI Use.}
AI tools were used as part of both the research methodology and manuscript
preparation. As described in the paper, OpenAI models and open-source models
were used for tasks including candidate verification, synthetic-data generation,
and model development. Microsoft Copilot and ChatGPT were also used to improve
the clarity and readability of the manuscript. All methodological decisions,
analyses, interpretations, and conclusions remain the responsibility of the
authors.

\textbf{Disclaimer.}
The findings, interpretations, and conclusions expressed in this paper are
entirely those of the authors. They do not necessarily represent the views of
the International Bank for Reconstruction and Development/World Bank and its
affiliated organizations, the Executive Directors of the World Bank, or the
governments they represent.

\bibliographystyle{iclr2025_conference}
\bibliography{iclr2025_conference}

\newpage

\appendix

\section{Appendix}
\label{sec:appendix}

\subsection{A. LLM-Assisted Annotation Refinement}
\label{app:llm_refinement}

This section provides additional implementation details for the LLM-assisted
annotation refinement described in Section~\ref{sec:methodology}. Candidate
dataset spans produced by the weakly supervised candidate generator are passed
to \texttt{GPT-4o} together with their surrounding context. The LLM determines
whether each candidate is a valid dataset reference and returns its decision
through OpenAI's structured outputs API.

The annotation rules are specified in two complementary ways. A Pydantic
response schema constrains the structure of the returned annotation, while a
system prompt defines what constitutes a valid dataset reference and the rules
used to distinguish datasets from other entities.

\subsubsection{Pydantic Response Format}

\begin{figure}[H]
\centering
\includegraphics[width=0.90\textwidth]{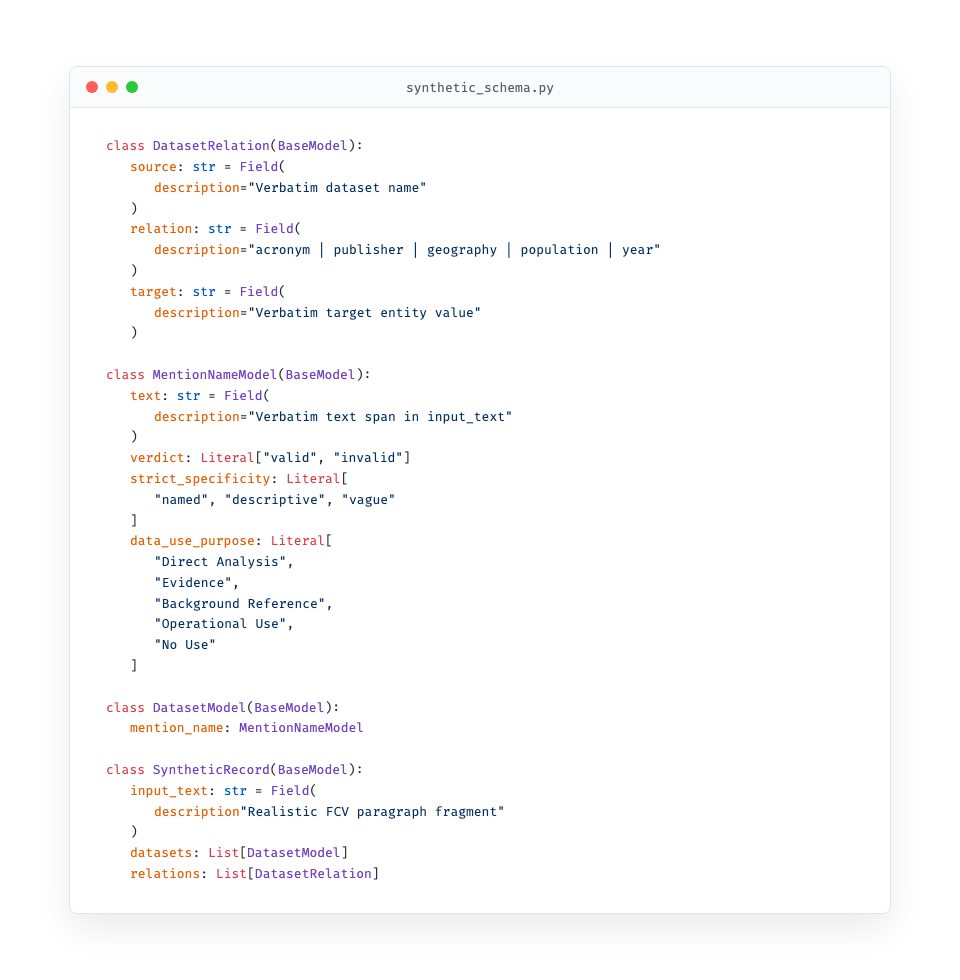}
\caption{Pydantic response schema used for LLM-assisted annotation refinement.}
\label{fig:pydantic_validation_schema}
\end{figure}

The response from the LLM judge is constrained using the Pydantic schema shown
in Figure~\ref{fig:pydantic_validation_schema}. The structured response format
ensures that validation decisions and associated annotations can be parsed
consistently and incorporated into the training-data construction pipeline.

\subsubsection{LLM Judge System Prompt}

The system prompt used to evaluate candidate dataset mentions is reproduced
below.

\begin{quote}
\small

\textbf{System Prompt}

You are an expert data annotator and LLM judge. Your task is to evaluate whether
a candidate text span extracted from a sentence is a valid ``Data Mention''
based on the strict guidelines below.

\textbf{Core Principle}

A valid data mention must refer directly to a structured data object or resource
(e.g., a database, survey, file, registry, microdata, census, or indicator
index).

\begin{itemize}
    \item Any citation, quote, or reference to data being used, relied upon, or
    quoted (e.g., in external reports, policy strategies, or cited literature)
    is a valid data mention.

    \item Vague or generic references to data resources (e.g., ``repeated survey
    data'', ``time-varying data on production'', ``microdata'', ``household
    data'', ``administrative records'') are valid vague data mentions.
\end{itemize}

\textbf{Exclusions (Flag as Invalid)}

\begin{enumerate}
    \item Author calculations and self-attributions (e.g., ``own elaboration'',
    ``SAG estimates'').

    \item Geographical or political entities on their own (e.g., ``Ghana'',
    ``Somalia'').

    \item Model configurations, statistical methodologies, or analytical
    frameworks (e.g., ``probit model'', ``sensitivity analysis'').

    \item Technical hardware, sensors, or datums (e.g., ``sensor'', ``WGS84'').

    \item Document structure and layout titles (e.g., ``Table 1'', ``Figure 3'',
    ``this report'').

    \item Qualitative research methodologies on their own (e.g., ``Focus Group
    Discussions'', ``Key Informant Interviews'').

    \item Bare organization names in opinion, actor, or decision roles
    (e.g., ``World Bank'', ``UNHCR'') unless combined with a data noun.

    \item Activity, process, or recommendation references (e.g., ``data
    collection and surveys'').

    \item Written reports, narrative publications, bulletins, policy briefs,
    and response plans (e.g., ``World Development Report 2022'',
    ``Humanitarian Response Plan'') when cited as background literature rather
    than an active structured database.

    \item PDF page-layout artifacts, running headers or footers, and repeating
    document or project titles (e.g., ``Costa Rica Results in Education
    (P181174)'', ``Zambia Agribusiness and Trade Project'').

    \item Table cells and structured form fields that merely list data sources
    without prose narrative. Only evaluate mentions that appear in complete
    prose sentences.
\end{enumerate}

\textbf{Strict Specificity Taxonomy}

\begin{itemize}
    \item \texttt{named}: The dataset has a proper name, acronym, or formal
    registry or branding title (e.g., ``Demographic and Health Survey'',
    ``DHS'', ``LSMS'').

    \item \texttt{descriptive}: The dataset is described in a specific way but
    does not use a proper name. It must contain at least two identifying
    details, such as who collected it, the population or geography covered,
    when it was collected, or how it was produced.

    \item \texttt{vague}: Generic data terms or variables without specific
    identifying details (e.g., ``survey data'', ``administrative records'').
\end{itemize}

\end{quote}

\subsection{B. Synthetic Data Generation}
\label{app:synthetic_generation}

Synthetic examples supplement the LLM-refined annotations obtained from real
documents. As described in Section~\ref{sec:methodology}, generation uses
few-shot prompting to produce additional examples of dataset references across
different domains and contexts. This section provides additional detail on the
domain variable banks, validation rules, and contrastive examples used in this
process.

\subsubsection{Domain Variable Banks}

To increase variation in the generated examples, the generation procedure draws
from six domain-specific banks containing dataset names and related data
resources. These banks steer generation toward terminology relevant to
development, humanitarian, and FCV applications.

\begin{table}[H]
\centering
\caption{Domain variable banks used to steer synthetic example generation.}
\label{tab:domain_banks}
\small
\begin{tabular}{p{0.25\textwidth} p{0.65\textwidth}}
\toprule
\textbf{Domain} & \textbf{Illustrative Data Resources and Terms} \\
\midrule

Development Economics &
Logistics Performance Index; CPIA; Enterprise Surveys; Doing Business;
Household Income and Expenditure Survey; LSMS; Poverty Assessment \\

Humanitarian \& Protection &
Multi-Sector Needs Assessment (MSNA); UNHCR Refugee Registry; Protection
Monitoring; mVAM; Displacement Tracking Matrix; Livelihoods Survey \\

Global Health &
DHS; Multiple Indicator Cluster Surveys (MICS); STEPwise approach to
surveillance; National Health Accounts; Malaria Indicator Survey \\

Food Security \& Agriculture &
Integrated Food Security Phase Classification (IPC); Comprehensive Food
Security and Vulnerability Analysis (CFSVA); Food Consumption Score \\

Education &
PISA; TIMSS; PIRLS; EGRA; EGMA; SACMEQ; Education Management Information
System (EMIS); learning assessment database \\

Climate \& Environment &
EM-DAT; global climate models; NOAA satellite data; IPCC emissions database;
water monitoring indicators \\

\bottomrule
\end{tabular}
\end{table}

\subsubsection{Validation of Synthetic Examples}

Generated examples are subjected to automated checks before inclusion in the
training corpus. These checks enforce three requirements:

\begin{enumerate}
    \item The generated passage must contain the requested dataset and associated
    metadata attributes verbatim.

    \item Extracted dataset spans must not include standalone publication years.

    \item Extracted spans must not terminate in incomplete grammatical fragments,
    such as trailing prepositions or verbs (e.g., ``of'', ``for'', or ``from'').
\end{enumerate}

Examples that fail these checks are excluded from the training corpus.

\subsubsection{Contrastive Examples}

Synthetic generation is also used to construct contrastive examples. These
examples are designed to distinguish dataset references from similar
expressions whose interpretation changes with context. The objective is to
reduce reliance on surface features such as capitalization, acronyms, or the
presence of terms such as ``survey.''

Table~\ref{tab:contrast_pairs_appendix} provides illustrative examples.

\begin{table}[H]
\centering
\caption{Illustrative contrastive examples used during synthetic-data
generation.}
\label{tab:contrast_pairs_appendix}
\small
\begin{tabular}{p{0.17\textwidth} p{0.46\textwidth}
p{0.20\textwidth} p{0.09\textwidth}}
\toprule
\textbf{Context} & \textbf{Example Sentence} &
\textbf{Target Span} & \textbf{Label} \\
\midrule

Dataset &
The analysis incorporates the \textbf{Living Standards Measurement Survey
(LSMS)} to calculate consumption aggregates. &
Living Standards Measurement Survey &
\texttt{named} \\

Non-dataset &
The \textbf{LSMS team} coordinated field operations and trained local
enumerators. &
None &
Empty \\

\midrule

Dataset &
We analyzed the \textbf{household consumption survey} data across three
displacement camps. &
household consumption survey &
\texttt{descriptive} \\

Non-dataset &
We conducted \textbf{household surveys} to assess immediate protection needs. &
None &
Empty \\

\bottomrule
\end{tabular}
\end{table}

The contrastive examples are intended to teach the model that the presence of a
familiar dataset name or data-related expression is not sufficient for
extraction. The surrounding context must indicate that the expression refers to
a data resource.

\subsection{C. Additional Evaluation Details}
\label{app:evaluation_details}

This section provides additional detail on the evaluation procedure described
in Section~\ref{sec:experimental_setup}. We evaluate the final fine-tuned model
at both the mention level and the passage level.

\subsubsection{Mention-Level Span Matching}

Mention-level evaluation compares each predicted dataset span with the
gold-standard annotation using token-level Jaccard similarity. For a predicted
span $S_p$ and gold span $S_g$, similarity is defined as

\begin{equation}
J(S_p,S_g)
=
\frac{|W_p \cap W_g|}
{|W_p \cup W_g|},
\end{equation}

where $W_p$ and $W_g$ denote the sets of tokens in the predicted and
gold-standard spans, respectively.

A prediction is considered a match when

\begin{equation}
J(S_p,S_g) \geq 0.5.
\end{equation}

This criterion allows minor differences in extraction boundaries while
requiring substantial overlap with the annotated dataset mention.

\subsubsection{Passage-Level Detection}

We additionally evaluate whether the model correctly determines if a text chunk
contains at least one dataset reference. A passage is classified as positive if
the model extracts one or more dataset mentions and negative if it returns no
extraction.

This provides a complementary evaluation of the model's ability to distinguish
dataset-bearing passages from passages containing no dataset reference, which
is important when applying the extractor to large document collections.

\subsection{D. Analysis of Filtered Candidate Mentions}
\label{app:rejected_candidates}

The weak-supervision stage deliberately uses a permissive prediction threshold
of $\tau=0.15$ to prioritize candidate recall. Consequently, the candidate
generator surfaces both potential dataset references and other entities that
share textual characteristics with dataset names. These predictions are
subsequently evaluated during LLM-assisted annotation refinement.

This behavior is particularly visible in operational documents, where project
titles, administrative systems, monitoring tools, document headings, and
information products often appear as capitalized multi-word expressions or
acronyms. To characterize these cases, Table~\ref{tab:invalid_candidates}
reports the 15 most frequent candidate strings filtered out during annotation
refinement in the FCV operational documents.

\begin{table}[H]
\centering
\caption{Top 15 most frequent candidate strings filtered out during LLM-assisted
annotation refinement in FCV operational documents.}
\label{tab:invalid_candidates}
\small
\begin{tabular}{r p{0.33\textwidth} c p{0.48\textwidth}}
\toprule
\textbf{Rank} & \textbf{Candidate String} & \textbf{Freq.} &
\textbf{Primary Reason for Exclusion} \\
\midrule

1 & Iraq COVID-19 Vaccination Project
& 25 & Project title, not a structured data resource. \\

2 & Formal Employment Creation Project
& 23 & Project title repeated as part of the document structure. \\

3 & Municipal Services Improvement Project...
& 18 & Project title, not a structured data resource. \\

4 & PROJECT FINANCING DATA
& 5 & Document structure header, not a dataset. \\

5 & INSTITUTIONAL DATA
& 4 & Section title within an operational reporting template. \\

6 & SYSTEMATIC OPERATIONS RISK-RATING TOOL
& 3 & Risk-assessment framework or tool, not a dataset. \\

7 & Procurement Risk Assessment \& Mgmt System
& 3 & Administrative system (PRAMS), not a dataset. \\

8 & Municipal Services Improvement Project
& 3 & Project name in an administrative context. \\

9 & Emergency Food Security Project
& 3 & Operational project title, not a structured data resource. \\

10 & Project Development Objective Indicators
& 2 & Monitoring framework terminology rather than an identifiable dataset. \\

11 & Systematic Tracking of Exchanges in Procurement
& 2 & Administrative procurement platform (STEP), not a dataset. \\

12 & TKYB ESMS
& 2 & Operational environmental and social management system. \\

13 & EU Needs Assessment
& 2 & Narrative publication or report rather than an underlying dataset. \\

14 & official monthly data
& 1 & Generic expression rejected under the annotation criteria in its
document context. \\

15 & Livelihoods Survey Findings
& 1 & Narrative publication or report rather than an underlying dataset. \\

\bottomrule
\end{tabular}
\end{table}

The filtered candidates are dominated by meaningful but out-of-scope entities
rather than arbitrary strings. In particular, project and program titles recur
frequently in operational documents and often have the same surface features as
named datasets. Administrative systems and monitoring tools present a similar
pattern. This illustrates why candidate generation and annotation refinement
serve different purposes in the framework: the first stage is designed to
surface plausible candidates broadly, while the second determines whether those
candidates actually refer to structured data resources.

\subsubsection{Illustrative Context Examples}

Candidate strings alone do not always provide enough information to determine
whether an expression refers to a dataset. Table~\ref{tab:context_examples}
illustrates how valid dataset references and filtered candidates appear in
context. The examples show why surrounding text is useful for distinguishing
structured data resources from projects, administrative platforms, qualitative
methods, and publications.

\begin{table}[H]
\centering
\caption{Illustrative contexts for valid dataset references and filtered
candidate mentions.}
\label{tab:context_examples}
\small
\begin{tabular}{p{0.15\textwidth} p{0.29\textwidth} p{0.49\textwidth}}
\toprule
\textbf{Status} & \textbf{Candidate Entity} &
\textbf{Context Sentence} \\
\midrule

\textbf{Valid} &
Demographic Health Survey &
``...estimates are derived using microdata from the
\textbf{Demographic Health Survey} conducted in the region.'' \\

\textbf{Valid} &
MSNA Poland 2023 survey &
``The \textbf{MSNA Poland 2023 survey} serves as the primary dataset for
profiling refugee household expenditure.'' \\

\textbf{Valid} &
ZUS data &
``We cross-reference these registration figures with monthly administrative
\textbf{ZUS data} on active social benefits.'' \\

\midrule

\textbf{Filtered} &
Iraq COVID-19 Vaccination Project &
``Under the \textbf{Iraq COVID-19 Vaccination Project}, the government
procured vaccines to support the immunization campaign.'' \\

\textbf{Filtered} &
Systematic Tracking of Exchanges in Procurement &
``All procurement processes must be logged in the
\textbf{Systematic Tracking of Exchanges in Procurement} platform.'' \\

\textbf{Filtered} &
Focus Group Discussions &
``Qualitative findings from the \textbf{Focus Group Discussions} are used to
supplement our statistical results.'' \\

\textbf{Filtered} &
Livelihoods Survey Findings &
``According to the \textbf{Livelihoods Survey Findings} published by the
agency, agricultural yields have declined.'' \\

\bottomrule
\end{tabular}
\end{table}

The examples reinforce the role of context-aware refinement. Terms such as
``Survey,'' ``Assessment,'' or an institutional acronym provide useful signals
during candidate generation, but they do not by themselves establish that the
expression refers to a dataset. The refinement stage instead evaluates the role
of the candidate in the surrounding text.

\end{document}